\documentclass[10pt,twocolumn,letterpaper]{article}

\usepackage{cvpr}
\usepackage{times}
\usepackage{epsfig}
\usepackage{graphicx}
\usepackage{epstopdf}
\usepackage{times}
\usepackage{helvet}
\usepackage{courier}
\usepackage{csquotes}
\usepackage{amsmath}
\usepackage{amssymb}
\usepackage{algorithm}
\usepackage{algorithmic}
\usepackage{multirow}

\usepackage[pagebackref=true,breaklinks=true,letterpaper=true,colorlinks,bookmarks=false]{hyperref}

\cvprfinalcopy % *** Uncomment this line for the final submission

\def\cvprPaperID{2293} % *** Enter the CVPR Paper ID here

\ifcvprfinal\fi
\begin{document}

%%%%%%%%% TITLE
\title{Accurate Single Stage Detector Using Recurrent Rolling Convolution}
\author{
Jimmy Ren \hspace{0.03in} Xiaohao Chen \hspace{0.03in} Jianbo Liu \hspace{0.03in} Wenxiu Sun \hspace{0.03in} Jiahao Pang \hspace{0.03in} Qiong Yan \hspace{0.03in} Yu-Wing Tai \hspace{0.03in} Li Xu\\\\
SenseTime Group Limited\\
\{rensijie, chenxiaohao, liujianbo, sunwenxiu, pangjiahao, yanqiong, yuwing, xuli\}@sensetime.com\\
}

\maketitle
\thispagestyle{empty}

%%%%%%%%% ABSTRACT
\begin{abstract}
Most of the recent successful methods in accurate object detection and localization used some variants of R-CNN style two stage Convolutional Neural Networks (CNN) where plausible regions were proposed in the first stage then followed by a second stage for decision refinement. Despite the simplicity of training and the efficiency in deployment, the single stage detection methods have not been as competitive when evaluated in benchmarks consider mAP for high IoU thresholds. In this paper, we proposed a novel single stage end-to-end trainable object detection network to overcome this limitation. We achieved this by introducing Recurrent Rolling Convolution (RRC) architecture over multi-scale feature maps to construct object classifiers and bounding box regressors which are ``deep in context''. We evaluated our method in the challenging KITTI dataset which measures methods under IoU threshold of 0.7. We showed that with RRC, a single reduced VGG-16 based model already significantly outperformed all the previously published results. At the time this paper was written our models ranked the first in KITTI car detection (the hard level), the first in cyclist detection and the second in pedestrian detection. These results were not reached by the previous single stage methods. The code is publicly available.\footnote{\url{https://github.com/xiaohaoChen/rrc_detection}}
\end{abstract}

%%%%%%%%% BODY TEXT
\section{Introduction}

In many real-world applications, robustly detecting objects with high localization accuracy, namely to predict the bounding box location with high Intersection over Union (IoU) to the groundtruth, is crucial to the quality of service. For instance, in vision based robotic arm applications, the process of generating robust and accurate operations in picking up an object are highly dependent on the object localization accuracy. In advanced driver assistance systems (ADAS), accurately localizing cars and pedestrians is also closely related to the safety of the autonomous actions.

\begin{figure}[t]
%\vskip 0.2in
\begin{center}
\centerline{\includegraphics[width=1\linewidth]{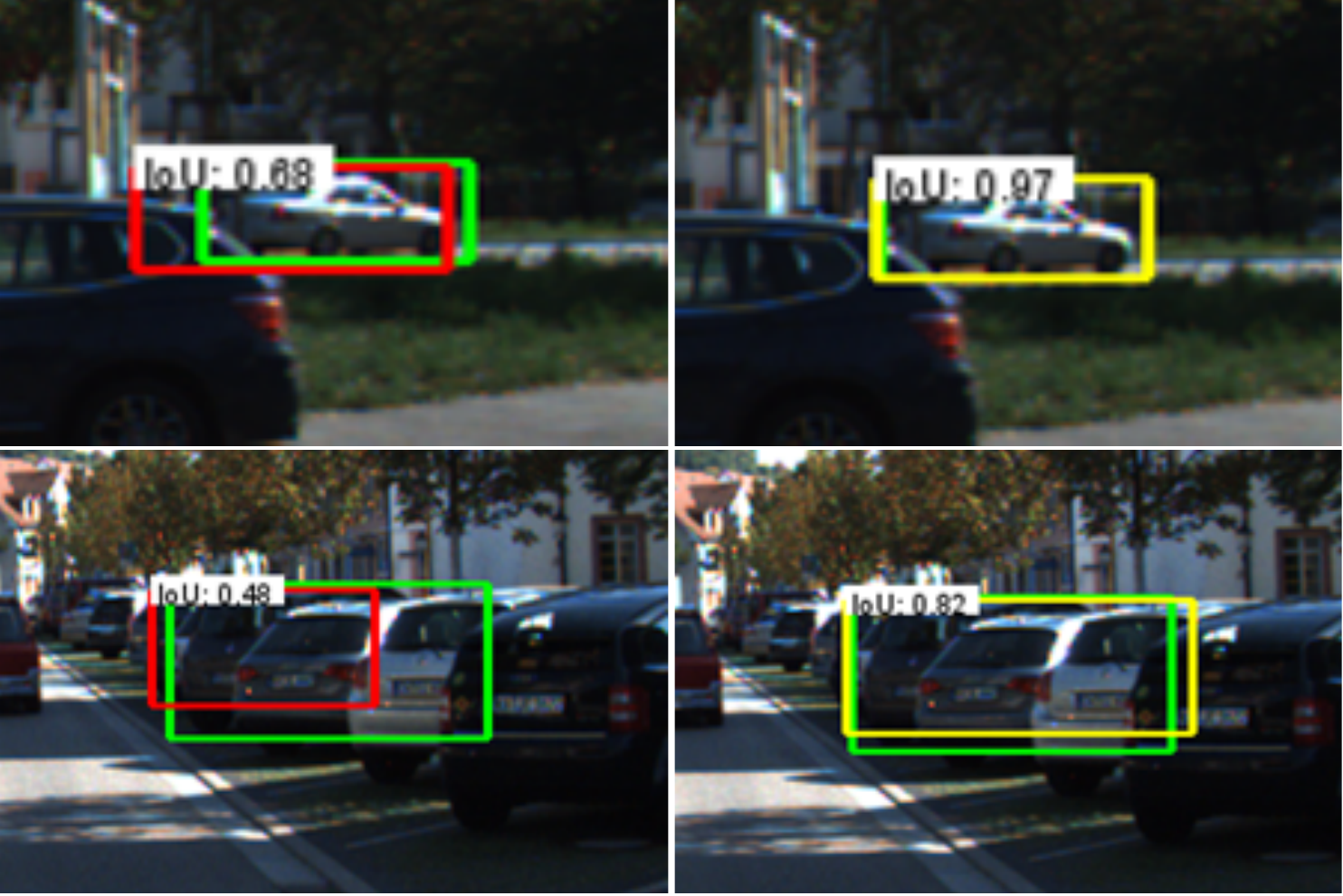}}
\caption{Left column: Previous single stage detector failed to generate bounding boxes of high IoU to the groundtruth bounding box (green) for small and occluded objects; Right column: With the proposed RRC, we can get high quality bounding boxes.}
\label{fig:fig1}
\end{center}
\vskip -0.2in
\end{figure}

Recent progress in object detection was heavily driven by the successful application of feed-forward deep Convolutional Neural Networks (CNN). Among many variants of the CNN based approaches, they can be roughly divded into two streams. The first is the R-CNN style \cite{girshick14CVPR} two stage methods. In these methods, plausible regions were proposed in the first stage then followed by a second stage for decision refinement. The other type of methods aimed to eliminate the region proposal stage and directly train a single stage end-to-end detector. The single stage detectors are usually easier to train and more computationally efficient in production \cite{liu15ssd}. However, such advantage is largely overwritten when the models are evaluated in benchmarks consider mAP for high IoU thresholds (e.g. KITTI car \cite{Geiger2012CVPR}) since the two stage methods are usually advantageous in performance. We will later show that this weakness of the single stage methods is not attribute to the inability in recognizing objects in complex scenes but the failure in generating high quality bounding boxes. Two examples are illustrated in the left column of figure \ref{fig:fig1}.

It can be experimentally shown that most of the low quality bounding boxes come from the failure localization of either small objects or overlapping objects. In either case, conventional bounding box regression becomes highly unreliable because the exact locations of the correct bounding boxes must be determined with the context (e.g. multi-scale information or feature around the occluded region). That is why it is effective to resort to some form of context aware refinement procedure to remove such errors. The RoI pooling and classification stage of Faster R-CNN can be thought of a simple method to take advantage of such context by resampling feature maps.

In this paper, we show that it is possible to seamlessly integrate the context aware refinement procedure in a single stage network. The insight is such procedure can be ``deep in context'' by using a novel Recurrent Rolling Convolution (RRC) architecture. In other words, contextual information can be gradually and selectively introduced to the bounding box regressor when needed. The whole process is fully data driven and can be trained end-to-end. We evaluated our method in the challenging KITTI dataset which considers mAP for high IoU thresholds. In our experiments, we used the reduced VGG-16 network instead of the full VGG network or the more recent ResNet as our pre-trained base network so that we are able to fully illustrate the effectiveness of the newly added RRC. This guarantees that such improvement is not simply introduced by the more powerful backbone network. The results showed that our approach significantly outperformed all the previously published results by a single model. An ensemble of our models ranks top among all the methods submitted to the benchmark.

The contributions of our work can be summarized as follows.
\vspace{-0.5\baselineskip}
\begin{itemize}
   \item First, we showed that it is possible to train a single stage detector in the end-to-end fashion to produce very accurate detection results for tasks requiring high localization quality.
   \vspace{-0.5\baselineskip}
   \item Second, we discovered that the key for improving single stage detector is to recurrently introduce context to the bounding box regression. This procedure can be efficiently implemented with the proposed Recurrent Rolling Convolution architecture.
\end{itemize}

\section{Related Work}

Convolutional neural network approaches with a region proposal stage have recently been very successful in the area of object detection. In the R-CNN paper \cite{girshick14CVPR}, selective search \cite{Uijlings13} was used to generate object proposals, CNN was used to extract and feed features to the classifier. Two acceleration approaches to R-CNN were later proposed. In \cite{girshickICCV15fastrcnn}, RoI pooling was used to efficiently generate features for object proposals. In \cite{RenSQ16}, the authors used CNN instead of selective search to perform region proposal. Many authors adopted the framework in \cite{RenSQ16} and proposed a number of variants which performs well in benchmarks consider mAP for high IoU threshold. For instance, in \cite{Yang16} the authors proposed to use scale-dependent pooling and layerwise cascaded rejection classifiers to increase the accuracy and obtained good results. Subcategory information was used in \cite{Xiang16} to enhance the region propose stage and achieved promising results in KITTI.

One problem with the R-CNN style methods is that in order to process a large number of proposals the computation in the second stage is usually heavy. Various single stage methods which do not rely on region proposals were proposed to accelerate the detection pipeline. SSD \cite{liu15ssd} is a single stage model in which the feature maps with different resolutions in the feed-forward process were directly used to detect objects with sizes of a specified range. This clever design saved considerable amount of computation and performed much faster than \cite{RenSQ16}. It achieved good results in datasets for IoU threshold of 0.5. However, we will show in our experiments that the performance drops significantly when we increase the bar for bounding box quality. YOLO \cite{Redmon16} is another fast single stage method which generated promising results, however, it's not as accurate as SSD though the customized version is faster. We noticed that fully convolutional two stage methods \cite{Dai16} has been proposed to reduce the computational complexity of the second stage. However, it heavily relies on the bigger and deeper backbone network. The motivation of \cite{Gidaris16} is similar to ours, but it does not consider contextual information by using recurrent architecture.

Though Recurrent Neural Networks (RNN) has been widely adopted in many areas such as image captioning \cite{Karpathy15,Xu15}, machine translation \cite{Sutskever14,Bahdanau15} and multimedia \cite{Ren16aaai}, the idea of using sequence modelling to improve object detection accuracy has been explored by only a few authors. An inspiring work is \cite{Stewart16} where the authors formalized the detection problem as a bounding box generation procedure and used Long Short-Term Memory (LSTM) \cite{Hochreiter97} to learn this procedure over deep CNN features by using the Hungarian loss. It was shown that this method is able to detect overlapping objects more robustly. However, in this formulation, the first bounding box in the sequence is essentially determined by a network ``shallow in context'' because the first output is only conditioned on the feature extracted by the last layer of the base network. This may be problematic if the first object in the pipeline is already challenging (e.g. small object, occluded, out of focus, motion blur, etc.) to detect which is not uncommon in many real-life applications. In addition, the method was only evaluated using IoU threshold of 0.5. Unlike \cite{Stewart16}, our proposed RRC architecture efficiently detects every object by a network which is ``deep in context'' and achieved state-of-the-art performance under a higher IoU threshold.

\section{Analysis and Our Approach}
\subsection{The Missing Piece of The Current Methods}
A robust object detection system must be able to simultaneously detect objects with drastically different scales and aspect ratios. In Faster R-CNN \cite{RenSQ16}, it relies on the large receptive field of each overlapping 3x3 area of the last convolutional layer to detect both small and large objects. Because multiple pooling layers are used, the resulting resolution of the last layer feature map is much smaller than the input image. This could be problematic for detecting small objects because in the low resolution feature map the features representing the fine details of the small objects is likely to be weak. Running the network over multi-scale input images as in \cite{Sermanet14} is one way to mitigate this issue but it is less computationally efficient.

An insightful alternative was proposed in the SSD paper \cite{liu15ssd}. This model exploits the fact that in most of the CNN models for detection, the internal feature maps in different layers are already of different scales due to pooling. Therefore, it is reasonable to utilize the higher resolution feature maps to detect relatively small objects and the lower resolution feature maps to detect relatively big objects. The advantage of this approach is that it not only provides an opportunity to localize the small objects more accurately by relocating the classification and bounding box regression of these objects to the higher resolution layers, as a single stage method it is also much faster than the previous two stage methods because such treatment for multi-scale does not add extra computation to the original backbone network.

However, SSD is not able to outperform state-of-the-art two stage methods. Actually, the gap becomes more significant when high IoU thresholds are used in evaluation. We now analyze and discuss why this is the limitation of SSD. We will also show how we addressed such limitation in our proposed single stage model and achieved state-of-the-art results in later sections. The utilization of multi-scale feature maps in SSD can be mathematically defined as follows,

\vspace{-0.5\baselineskip}
\begin{eqnarray}
\Phi_{n} = f_{n}(\Phi_{n-1}) = f_{n}(f_{n-1}(...f_{1}(I))), \label{eq:std_ssd1} \\
Detection = \mathcal{D}(\tau_{n}(\Phi_{n}),...,\tau_{n-k}(\Phi_{n-k})), n \! > \! k \! > \! 0,
\label{eq:std_ssd2}
\end{eqnarray}

\noindent where $\Phi_{n}$ is the feature maps in the layer $n$, $f_{n}(\cdot)$ is the nonlinear block to transform the feature maps in the $(n-1)th$ layer to the $nth$ layer. $f_{n}(\cdot)$ could be the combination of convolutional layers, pooling layers, ReLU layers, etc., $f_{1}(I)$ is the first nonlinear block to transfer the input image $I$ to the first layer feature maps. $\tau_{n}(\cdot)$ is the function to transform the $nth$ layer feature maps to the detection results for a certain scale range. $\mathcal{D}$ is the final operation to aggregate all the intermediate results and generate the final detection.

According to eq. (\ref{eq:std_ssd2}), we can find that it heavily relies on a strong assumption to perform well. Because the feature maps in each layer is solely responsible for the output of its scale, the assumption is that every $\Phi$, by itself, has to be sophisticated enough to support the detection and the accurate localization of the objects of interest. By sophistication it means that 1) the feature map should have enough resolution to represent the fine details of the object; 2) the function to transform the input image to the feature maps should be deep enough so that the proper high level abstraction of the object is built-in to the feature maps; 3) the feature maps contain appropriate contextual information based on which the exact location of the overlapping objects, occluded objects, small objects, blur or saturated objects can be inferred robustly \cite{RenSQ16,liu15ssd,Stewart16}. From eq. (\ref{eq:std_ssd1}) and (\ref{eq:std_ssd2}), we observed that $\Phi_{n}$ is much deeper than $\Phi_{n-k}$ when $k$ is large, so the aforementioned second condition does not hold for $\Phi_{n-k}$. The consequence is that $\tau_{n-k}(\cdot)$, the function to transform the feature maps in the $(n-k)th$ layer to its detection output, is likely to be a lot weaker and significantly harder to train than $\tau_{n}(\cdot)$. Faster R-CNN does not have this depth problem because its region proposals are generated from the last layer feature maps, namely

\vspace{-0.5\baselineskip}
\begin{equation}
Region \; proposals = \mathcal{R}(\tau_{n}(\Phi_{n})), n > 0.
\label{eq:rpn}
\end{equation}

\noindent However, eq. (\ref{eq:rpn}) also has its own problem because it does break the first condition. Therefore, we argue that a more reasonable function to learn in a single stage detector can be defined as follows

\vspace{-0.5\baselineskip}
\begin{equation}
\begin{split}
Detection = \mathcal{\hat{D}}(\tau_{n}(\hat{\Phi}_{n}(\mathcal{H})),\tau_{n-1}(\hat{\Phi}_{n-1}(\mathcal{H})),\\
...,\tau_{n-k}(\hat{\Phi}_{n-k}(\mathcal{H}))), \\ 
\mathcal{H} = \{\Phi_{n}, \Phi_{n-1}, ..., \Phi_{n-k}\}, n \! > \! k \! > \! 0, \\
size(\Phi_{n-k}) = size(\hat{\Phi}_{n-k}(\mathcal{H})), \forall k
\end{split}
\label{eq:good_ssd}
\end{equation}

\begin{figure*}[ht]
%\vskip 0.2in
\begin{center}
\centerline{\includegraphics[width=1.05\linewidth]{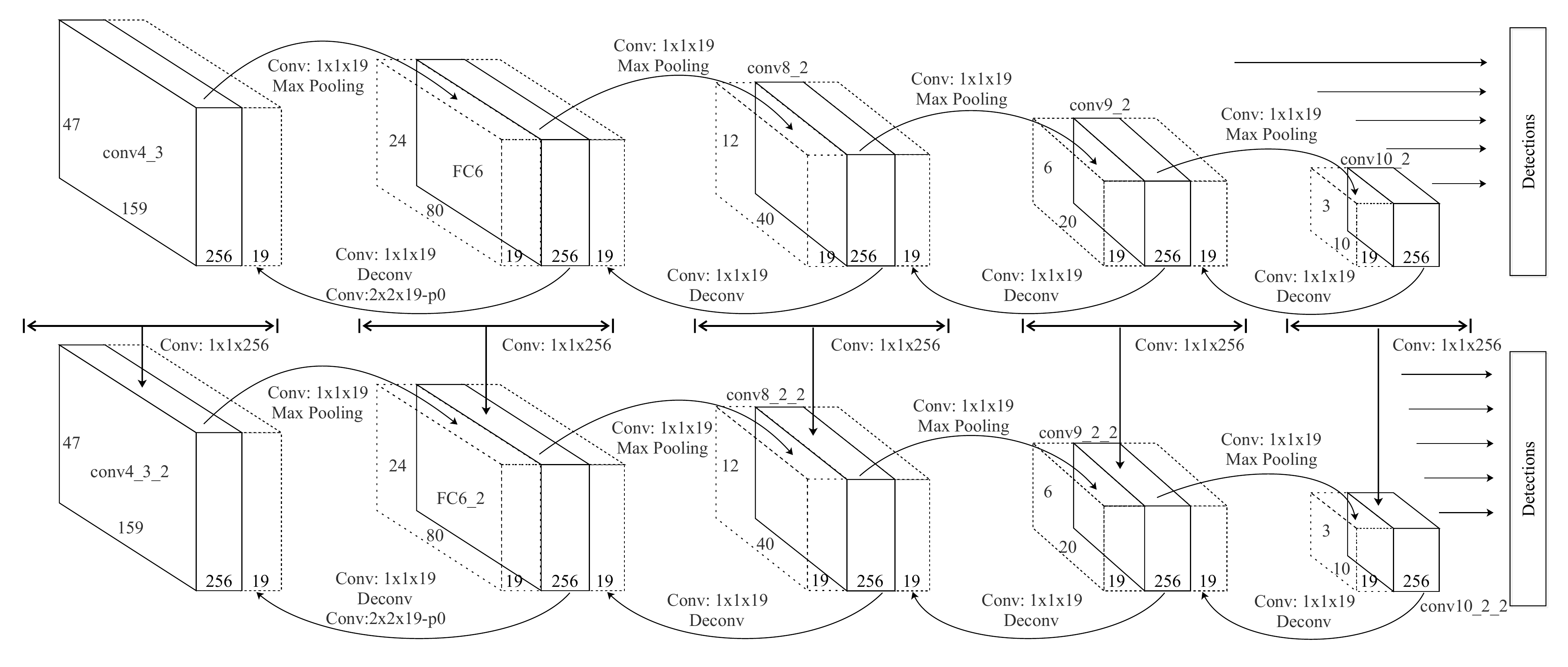}}
\caption{The Recurrent Rolling Convolution architecture. The diagram illustrates RRC for two consecutive iterations. All the feature maps (solid boxes) in the first stage including $conv4\_3$, $FC6$, $conv8\_2$, $conv9\_2$ and $conv10\_2$ were previously computed by the backbone reduced VGG16 network. In each stage, the arrows illustrates the top-down/bottom-up feature aggregation. All the weights of such feature aggregation are shared across stages. The selected features by the arrows are concatenated to the neighboring feature maps and illustrated by the dotted boxes. Between the stages, there are additional 1x1 convolution operators to transform the aggregated feature maps to their original sizes so that they are ready for the next RRC. These weights are also shared across iterations. Each RRC iteration has its own outputs and also connects to its own loss functions during training.}
\label{fig:rrc}
\end{center}
\vskip -0.2in
\end{figure*} 

\noindent where $\mathcal{H}$ is a set which contains all the feature maps contribute to the detection function $\mathcal{D}(\cdot)$ in eq. (\ref{eq:std_ssd2}). Unlike in eq. (\ref{eq:std_ssd2}), $\hat{\Phi}_n(\cdot)$ is now a function in which all the contributing feature maps are considered and outputs a new feature representation of the same dimensionality to $\Phi_n$.

The function $\mathcal{\hat{D}}(\cdot)$ defined in eq. (\ref{eq:good_ssd}) does satisfy the first two conditions of feature map sophistication because the feature maps outputted by $\hat{\Phi}_{n-k}(\mathcal{H})$ not only share the same resolution as $\Phi_{n-k}$, but also incorporate the features extracted in the deeper layers. It is worth noting that $\mathcal{\hat{D}}(\cdot)$ is still a single stage process though the modification to eq. (\ref{eq:std_ssd2}). In other word, if we can also make eq. (\ref{eq:good_ssd}) satisfy the third aforementioned condition and devise an efficient architecture to train it, we will be able to comprehensively overcome the limitations of the previous single stage methods and have the opportunity to surpass the two stage methods even for high IoU thresholds.

\subsection{Recurrent Rolling Convolution}
\paragraph{RNN for Conditional Feature Aggregation}
We now define details in $\hat{\Phi}(\mathcal{H})$ so that the feature maps generated by this function contains useful contextual information for detection. The contextual information in $\hat{\Phi}(\cdot)$ means differently for different objects of interest. For instance, when detecting small objects it means $\hat{\Phi}(\cdot)$ should return feature maps contain higher resolution features of this object to represent the missing details. When detecting occluded objects, $\hat{\Phi}(\cdot)$ should return feature maps contain robust abstraction of such object so that the feature is relatively invariant to occlusion. When detecting overlapping objects, $\hat{\Phi}(\cdot)$ should return feature maps contain both the details of the boundary and the high level abstraction to distinguish different objects. Nevertheless, for an intermediate level feature map such as $\Phi_{p}$ where $p$ is a positive integer, all the aforementioned contextual information can be retrieved either from its lower level counterparts $\Phi_{p-q}$ or its higher level counterparts $\Phi_{p+r}$, where $q$ and $r$ are also positive integers. The difficulty is that it is very hard to manually define a fixed rule for the function $\hat{\Phi}_{p}(\mathcal{H})$ to retrieve the appropriate features from $\Phi_{p-q}$ and $\Phi_{p+r}$ in $\mathcal{H}$, it is also very hard to manually select $q$ and $r$. Therefore, we must systematically learn this feature retrieval and aggregation process from the data.

However, the learning of $\hat{\Phi}(\mathcal{H})$ could be troublesome because $\mathcal{H}$ is a set containing multiple feature maps in different layers and of  different scales and we do not know which one should be involved and what kind of operations should be imposed to the feature map for the current object of interest. Therefore, a direct mapping from $\mathcal{H}$ to a useful $\hat{\Phi}(\mathcal{H})$ have to resort to a considerable size deep network with multiple layers of nonlinearity. This will not make a computationally efficient and easy to train single stage network. The alternative is to design an iterative procedure in which each step makes a small but meaningful and consistent progress. This procedure can be mathematically described as follows,

\vspace{-0.5\baselineskip}
\begin{equation}
\begin{split}
\hat{\Phi}_p^{t+1} = \mathcal{F}(\hat{\Phi}_p^t, \hat{\Phi}_{p-1}^t, \hat{\Phi}_{p+1}^t;\mathcal{W}), t > 0, \\
\hat{\Phi}_n^t = \Phi_n,\forall n \; when \; t = 1,
\end{split}
\label{eq:rnn_refine}
\end{equation}

\noindent where $\mathcal{F}$ is a function maps only $\hat{\Phi}_p^t$ and its direct higher and lower level counterparts at step $t$ to a new $\hat{\Phi}_p$ at step $t+1$. The function $\mathcal{F}$ is parametrized by some trainable weights $\mathcal{W}$.

\begin{figure}[ht]
%\vskip 0.2in
\begin{center}
\centerline{\includegraphics[width=0.9\linewidth]{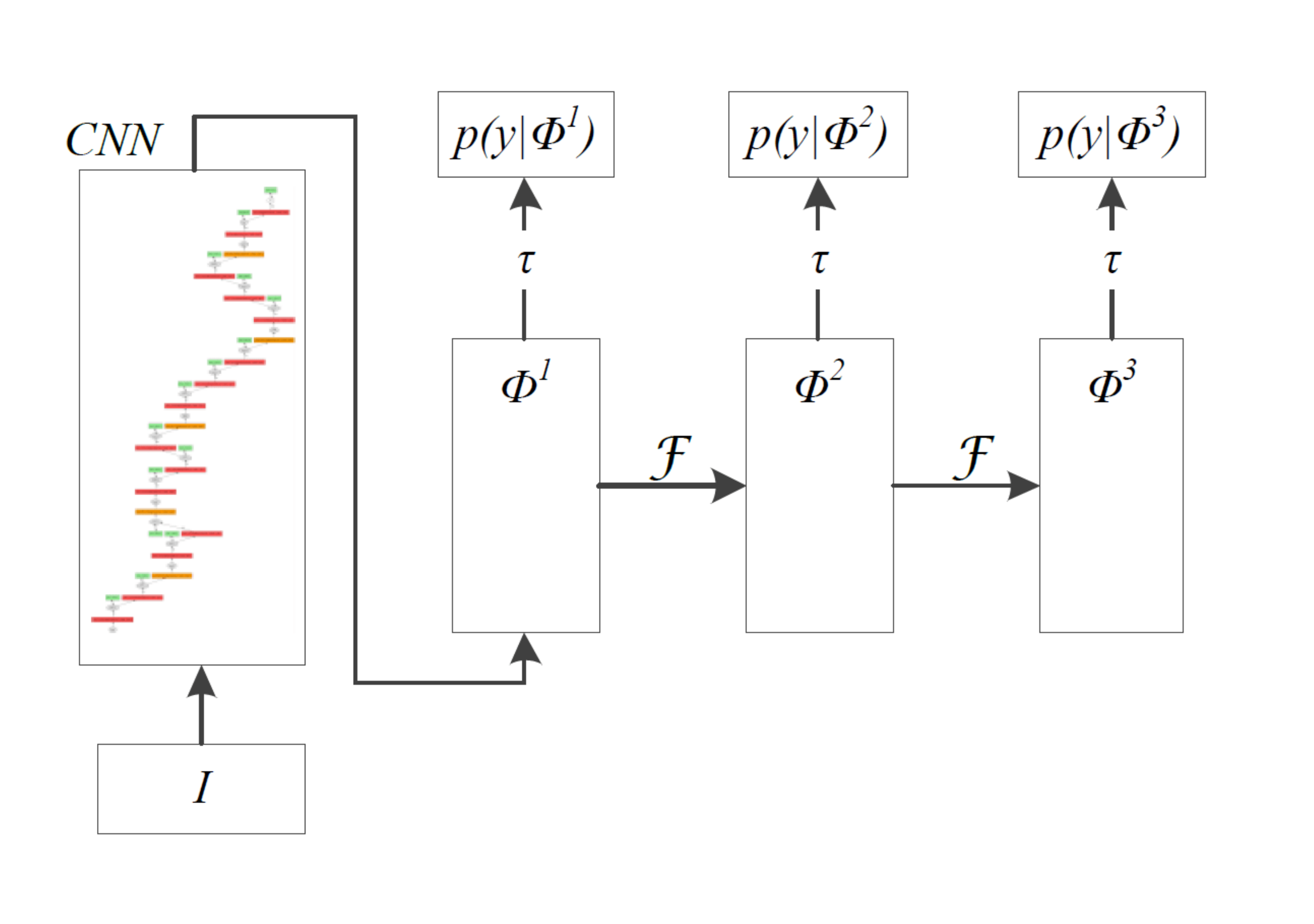}}
\caption{Illustration of recurrent feature aggregation.}
\label{fig:rnn1}
\end{center}
\vskip -0.2in
\end{figure} 

The equation is pictorially illustrated in figure \ref{fig:rnn1}. We can see from the figure that $I$ is the input image which is fed to the network and outputs the feature map $\hat{\Phi}^1$. When the function $\tau$ is applied to it for classification and bounding box regression, the output is only conditioned on $\hat{\Phi}^1$. Then the function $\mathcal{F}$ shall perform the feature aggregation to bring necessary contextual information and give a new $\hat{\Phi}^2$ at step 2. Then the function $\tau$ is able to output a refined result which is conditioned on the updated feature map $\hat{\Phi}^2$. Note that we can impose a supervision signal to each step during training so that the system finds useful contextual information in the feature aggregation to make real progress in detection. An important insight is that if the weights in $\mathcal{F}$ and $\tau$ are shared over steps respectively, this is a recurrent network. Recurrence can not be overlooked here because it ensures the consistent feature aggregation across the steps. This makes the feature aggregation in each step smooth and generalize well. Otherwise, it will be more prone to overfitting and cause unexpected bias.

\paragraph{RRC Model Details}
If we simultaneously apply eq. (\ref{eq:rnn_refine}) to every $\hat{\Phi}$, this is our proposed Recurrent Rolling Convolution model. It is worth noting that even though $\hat{\Phi}_p^{t+1}$ is a function of $\hat{\Phi}_p^{t}$ and its direct counterparts $\hat{\Phi}_{p-1}^{t}$ and $\hat{\Phi}_{p+1}^{t}$, if there are separate $\mathcal{F}$ for $\hat{\Phi}_{p-1}^{t}$ and $\hat{\Phi}_{p+1}^{t}$ respectively for their own direct counterparts, the values in $\hat{\Phi}_p^{t+1}$ will eventually be influenced by all the feature maps in $\mathcal{H}$ after enough iterations. 

The proposed RRC model is illustrated in figure \ref{fig:rrc} in detail. The figure shows how we applied RRC to the KITTI dataset using the reduced VGG-16 backbone model \cite{liu15ssd,liu15}. The size of the input images is 1272x375 with 3 channels, thus the sizes of the original $conv4\_3$ layer and $FC7$ layer are 159x47x512 and 80x24x1024 respectively, where 512 and 1024 are channel numbers. We used additional 3x3 convolutional layers to further reduce the channels of them to 256 before feature aggregation. Following SSD, we also used the layer $conv8\_2$, $conv9\_2$ and $conv10\_2$ for multi-scale detection, the difference is that our $conv8\_2$ layer has 256 instead of 512 channels. We found the unified channel number among multi-scale feature maps promotes more consistent feature aggregation.

We used one convolution layer and one deconvolution layer to aggregate features downwards. For instance, for the layer $conv8\_2$ a convolution layer with 1x1 kernel is used to generate feature maps of size 40x12x19. They are concatenated to $FC7$ after going through a ReLU and a deconvolution layer. Likewise, all the left pointing arrows in the figure indicate such downwards operations. We used one convolution layer and one max pooling layer to perform upwards feature aggregation. Also take the layer $conv8\_2$ as an example, a 1x1 convolution is followed by ReLU and max pooling, the resulting 20x6x19 feature maps are concatenated to $conv9\_2$. Similarly, all the right pointing arrows in the figure indicate such upwards operations. We call this feature aggregation procedure ``rolling'' because the left pointing and the right pointing arrows resemble it.

Once the rolling is done for the first time, 1x1 convolution is performed for each layer respectively to reduce the number of channels to the original setting. After this channel reduction, the whole feature aggregation is done for the first iteration. This channel reduction is important because it ensures a unified shape for every feature map between the two consecutive feature aggregation. It also makes the recurrent rolling possible. During training, the convolution kernels corresponding to each arrow as well as the channel reduction are all shared across iterations. We call this iterative process recurrent rolling convolution.

\paragraph{RRC Discussion}
RRC is a recurrent process in which each iteration gathers and aggregates relevant features for detection. As we discussed before, these revevant feature contains contextual information which is critical for detecting challenging objects. For each RRC, there is a separate loss function to guide the learning of it. This makes sure that relavant features will be gradually imported and makes the real progress we expect in every iteration. Because RRC can be performed multiple times, the resulting feature maps is therefore ``deep in context''. Different from \cite{Stewart16}, because RRC is not tailored for any particular bounding box therefore the depth in contextual information can be utilized to detect every object in the scene.

\paragraph{Loss Functions}
Each iteration has its own loss functions during training. Following SSD, the loss function for object category classification was cross-entropy loss. Smooth L1 loss was used for bounding box regression.

\paragraph{Bounding Box Regression Space Discretization}
In our setting, a group of feature maps in a layer (e.g. $conv4\_3$) is responsible for the regression for bounding boxes of a certain size range. Because the bounding box regression is essentially a linear process, thus if this range is too large or the feature is too complex, the robustness of the bounding box regression shall be significantly affected. Because the RRC process brings more contextual information to the feature maps, it will inevitably make the feature maps richer based on which the bounding box regression could be harder to do for the original object range. To overcome this issue and make the bounding box regression more robust, we further discretize the bounding box regression space within a particular feature maps by assigning multiple regressors for it so that each regressor is responsible for an easier task.

\begin{figure}
  \centering
  \includegraphics[width=1\linewidth]{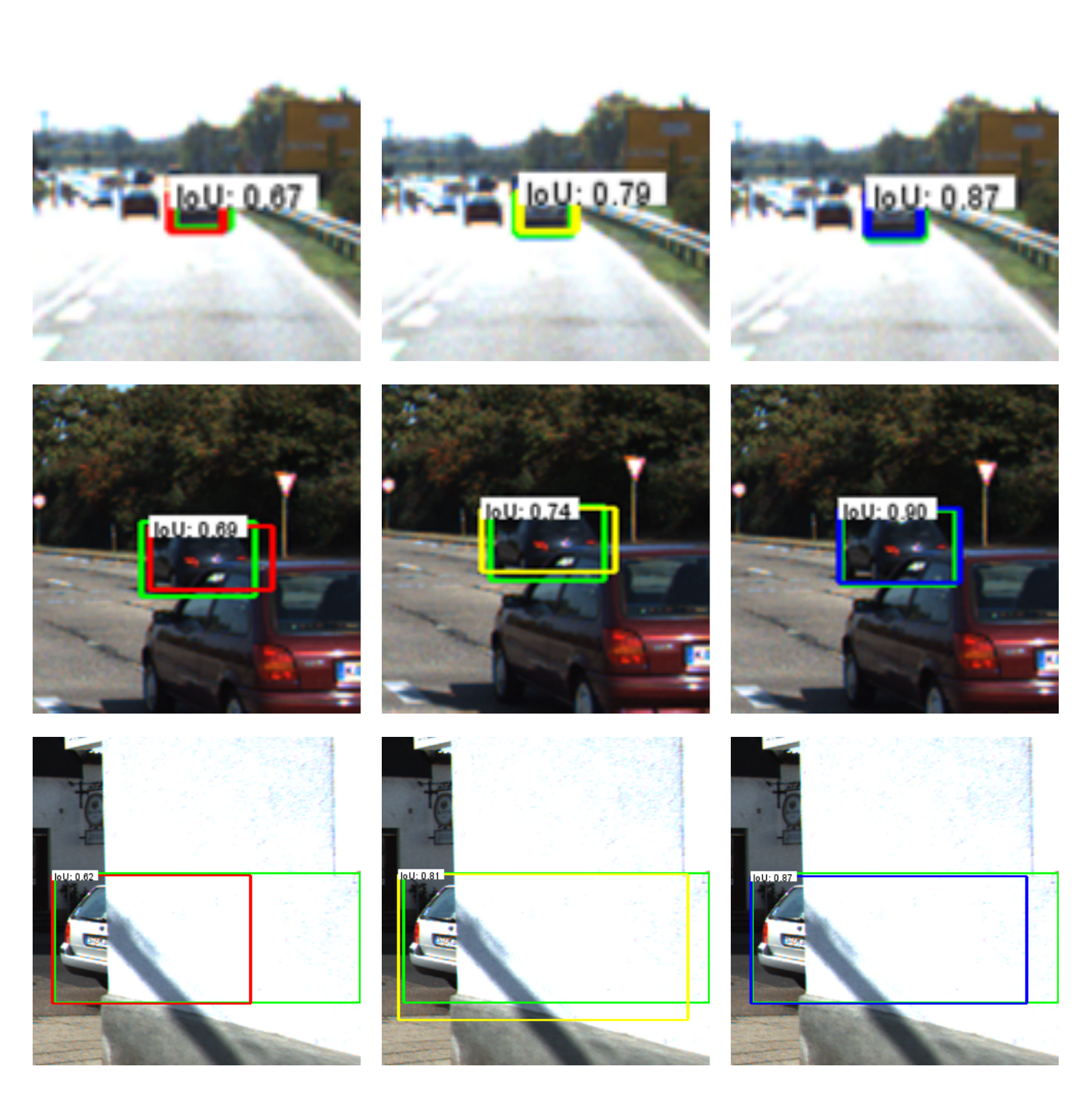}
  \caption{Comparison between SSD and RRC. Left column: results of SSD, failed to generate bounding box with IoU bigger than 0.7 to the groundtruth; Middle column: RRC, NMS over output 2 through output 6; Right column: RRC, NMS over output 3 through output 5. }
  \label{fig:evaluation1}
  \vskip -0.2in
\end{figure}

\section{Experiments}
The evaluation of our model was performed on the KITTI benchmark \cite{Geiger2012CVPR} which not only contains many challenging objects such as small and severely occluded cars and pedestrians, it also adopts an IoU threshold of 0.7 for the evaluation in the car benchmark. The KITTI dataset contains 7481 images for training and validation, and another 7518 images for testing. We did not use any other dataset in our experiments to enhance the results. The groundtruth of the test set is not publicly available. One needs to submit the results to a dedicated server for the performance evaluation of the test set.

We conducted three experiments in this paper. The first experiment examined the quality of the predictions after each recurrent rolling convolution. The second one evaluated the performance of our method in a smaller validation set. The final one evaluated our method in the official test set and compared with other state-of-the-art methods.

\begin{table}[!hbp] 
\centering
\caption{Average Loss of Different Predictions}
\label{table:rrc_outputs}
\begin{tabular}{|c||c|c|}
\hline
\hline
Output index & Training set & Validation set  \\
\hline
1& 0.662 & 1.461\\
%\hline
2& 0.622 & 1.374\\
%\hline
3& 0.609& 1.357\\
%\hline
4& 0.607& 1.361\\
%\hline
5& 0.609& 1.366\\
%\hline
6& 0.617& 1.375\\
\hline
\end{tabular}
\end{table}

\paragraph{Implementation Details}

The following settings were used throughout the experiments. For the network architecture, we did RRC for 5 times in training. We assigned 5 separate regressors for each corresponding feature map. Because RRC is performed by 1x1 convolutions, the resulting model is efficient. For data augmentation, in addition to the data augmentation methods adopted in the SSD paper we also randomly adjusted the exposure and saturation of the images by a factor of 1.3 in the HSV color space. In addition, as the minimum scale of the objects in the KITTI dataset is much smaller than the original configuration, we adjusted the corresponding scale of $conv4\_3$ from 0.1 to 0.066. We also removed the last global pooling layer of the original SSD model and set the scale of $conv10\_2$ to 0.85. For learning, stochastic gradient descent (SGD) with momentum of 0.9 was used for optimization. Weight decay was set to 0.0005. We set the initial learning rate to 0.0005. The learning rate will be divided by 10 every 40,000 iterations. We also adopted a simple image similarity metric for training set and validation set separation. The goal was to make the training set as different from the validation set as possible. Our resulting validation set has 2741 images.

\subsection{Examining The Outputs After Each RRC}
Because RRC was used for 5 times in the training, in principle our model has 6 outputs, namely the model makes 6 consecutive predictions. According to the design of RRC, we should be able to observe improvements after each RRC. The purpose of this experiment is to examine whether this is indeed the case.

To see the results, we ran a RRC model on both the training set and the validation set to calculate the average loss for both sets. The results are summarized in table \ref{table:rrc_outputs}.

The first output is the one before any RRC occurs. The second prediction happens after the first RRC iteration and so forth. We can see that the validation loss is generally bigger than the training loss. This indicates a certain degree of overfitting. This is normal because we reserved a significant portion of the images for the validation set. We observed a consistent trend in the table. The loss of the second output is significantly lower than the first one. The lowest loss is from the third or the fourth output. However, the ensuing loss values stop to decrease. 

In general, the results showed that RRC is able to significantly and consistently improve the predictions for a few consecutive recurrent iterations, but the improvement diverges eventually. The insights of this phenomenon are two-fold. First, the effectiveness of RRC can be confirmed. In later experiments, we will show that such improvement is the key for our proposed method to achieve state-of-the-art performance. On the other hand, the reason why RRC eventually degenerates the prediction is mainly because the lack of an effective memory mechanism \cite{Hochreiter97} which is a way to model the patterns in a long term sequence. Though a memory mechanism is helpful, it will also introduce considerable extra computation and memory cost to the model. We leave the design of an efficient memory mechanism for detection to our future work. This experiment provided a guidance for us on prediction selection for the final results. For the rest of the experiments, the final detection results were generated by running non-maximum suppression (NMS) over the third, fourth and the fifth predictions.

\subsection{Performance Evaluation on The Validation Set}

\begin{table} \scriptsize
\centering
\caption{Results on the KITTI validation set for different IoU thresholds}
\label{table:ssd_compare}
\begin{tabular}{|c|c|c|c|c|c|}
\hline
\hline
Methods & 0.6 &0.65& 0.7 & 0.75&0.8 \\
\hline
SSD&$90.43\%$&$90.15\%$&$89.16\%$&$87.22\%$&$76.12\%$\\
RRC*&$94.33\%$&$90.78\%$&$90.32\%$&$89.01\%$&$82.19\%$\\
RRC&$94.52\%$&$90.81\%$&$90.65\%$&$89.27\%$&$82.82\%$\\
\hline
\end{tabular}
\end{table}

\begin{table}
\centering
\caption{Results on the KITTI Car testing set (moderate)}
\label{table:final_compare}
\begin{tabular}{|c||c|}
\hline
\hline
\multirow{2}{*}{Methods} & \multicolumn{1}{c|}{Car}  \\
 &Moderate\\
\hline
SubCNN \cite{Xiang16}&$89.04\%$\\
MS-CNN \cite{Cai16}&$89.02\%$\\
SDP+RPN \cite{Yang16}&$88.85\%$\\
Mono3D \cite{Chen16cvpr}&$88.66\%$\\
3DOP \cite{Chen15nips}&$88.64\%$\\
\hline
RRC(single)&$89.85\%$\\
RRC(ensemble)&$90.19\%$\\
%\belowspace
\hline
\end{tabular}
\end{table}

\begin{table}
\centering
\caption{Results on the KITTI Car testing set (hard)}
\label{table:final_compare2}
\begin{tabular}{|c||c|}
\hline
\hline
\multirow{2}{*}{Methods} & \multicolumn{1}{c|}{Car}\\
 &Hard\\
\hline
DuEye (anonymous)&$86.18\%$\\
Genome (anonymous)&$85.82\%$\\
eagle (anonymous)&$85.66\%$\\
RV-CNN (anonymous)&$85.43\%$\\
\hline
RRC (ours)&$86.97\%$\\
%\belowspace
\hline
\end{tabular}
\end{table}

\begin{table}
\centering
\caption{Results on the KITTI Pedestrian testing set (moderate)}
\label{table:final_compare3}
\begin{tabular}{|c||c|}
\hline
\hline
\multirow{2}{*}{Methods} & \multicolumn{1}{c|}{Pedestrian}  \\
 &Moderate\\
\hline
SubCNN \cite{Xiang16}&$73.70\%$\\
MS-CNN \cite{Cai16}&$71.33\%$\\
SDP+RPN \cite{Yang16}&$70.16\%$\\
\hline
RRC (ours)&$75.33\%$\\
%\belowspace
\hline
\end{tabular}
\end{table}

\begin{table}
\centering
\caption{Results on the KITTI Cyclist testing set (moderate)}
\label{table:final_compare4}
\begin{tabular}{|c||c|}
\hline
\hline
\multirow{2}{*}{Methods} & \multicolumn{1}{c|}{Cyclist}  \\
 &Moderate\\
\hline
SubCNN \cite{Xiang16}&$71.06\%$\\
MS-CNN \cite{Cai16}&$75.46\%$\\
SDP+RPN \cite{Yang16}&$73.74\%$\\
\hline
RRC (ours)&$76.47\%$\\
%\belowspace
\hline
\end{tabular}
\end{table}

In this section, we quantify the improvement our method brings to the detection task. This experiment was carried out with the car dataset because the separation of training set and validation set for pedestrian is difficult due to the scarce data.

The mAP with different IoU thresholds was adopted for evaluation. We carefully trained a SSD car detector as the baseline. Two settings of RRC was tested in the experiment. The first one performed NMS over the output 2 through output 6 and denoted as RRC*. The second one adopted the standard approach which only performed NMS over the output 3 through output 5. As shown in table \ref{table:ssd_compare}, both of the RRC detectors performed much better than the original SSD implementation. Because both SSD and our method share the same reduced VGG-16 network as the backbone network, the improvement was the direct consequence of the newly added RRC architecture.

We can also observe that RRC performed consistently better than RRC*. This double confirmed the robustness of the conclusion we obtained from the first experiment. It is worth noting that if we increase the bar of IoU threshold to 0.8, RRC outperformed the original SSD by more than 6\%. This strongly confirmed the effectiveness of RRC in predicting high quality bounding boxes. Figure \ref{fig:evaluation1} illustrated some examples of the difference between the three models. Actually, the illustrated issue with SSD is not uncommon, the quality of the bounding box is a bottleneck to prevent the previous single stage methods from achieving good results. The issue can be well addressed by RRC.

\begin{figure*}[thb!]
  \includegraphics[width=1.05\linewidth]{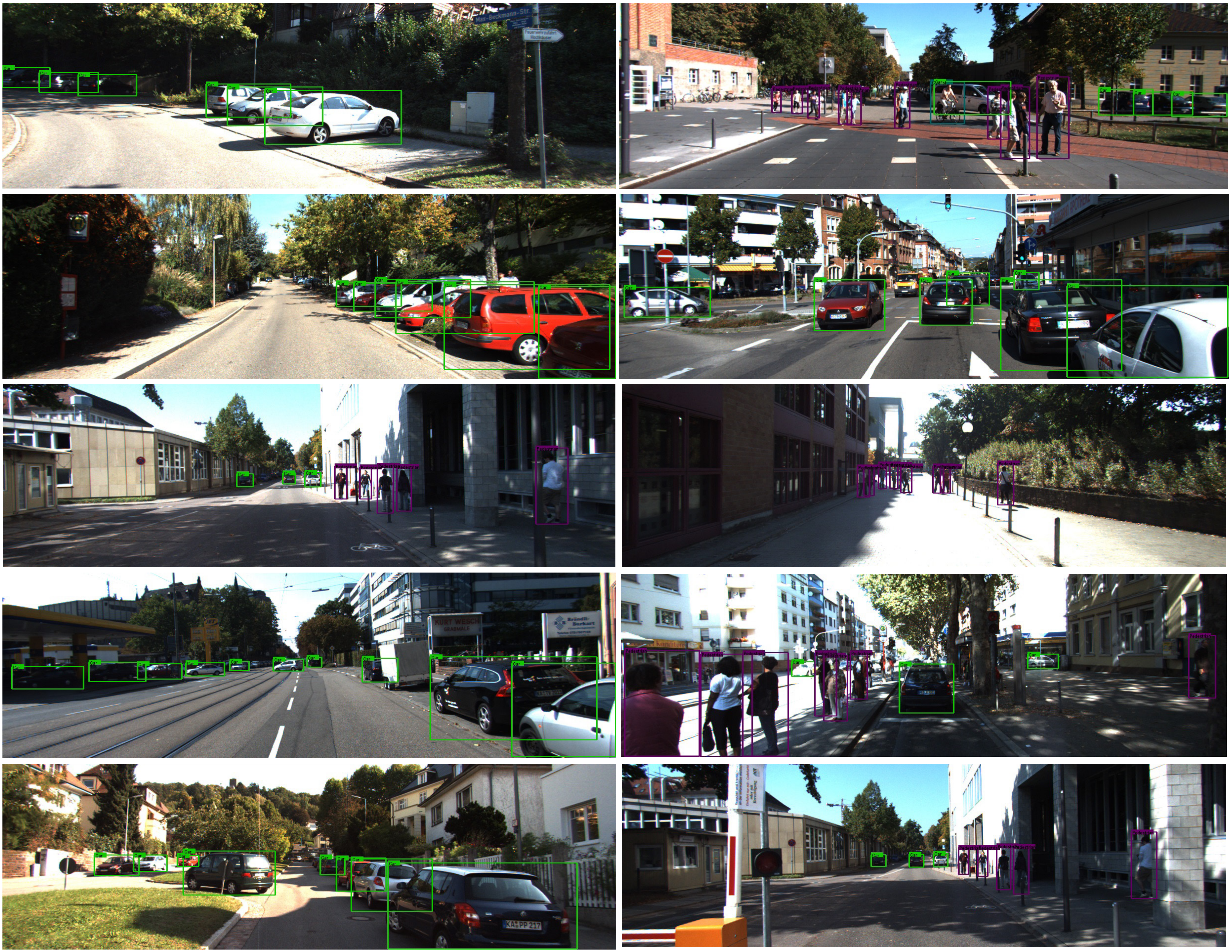}
  \caption{Detection results of our method in KITTI testing set.}
  \label{fig:final}
\end{figure*}

\subsection{Performance Evaluation on The Test Set}

In this experiment, we evaluated our model in the KITTI official test set for car, pedestrian and cyclist. For the car benchmark, we compared our method with five recently published state-of-the-art methods. The results are showed in table \ref{table:final_compare}. The only difference from the previous experiments is that we adopted the same input resolution as in \cite{Cai16} to ensure fair comparison. We can see that a single RRC model was able to significantly outperform all the previous published results. An ensemble of the RRC models performed much better than the previous methods. We also compared RRC with other anonymous unpublished submission to KITTI in table \ref{table:final_compare2}. By the time this paper was written, our results for the hardest category ranked the first among all the submitted methods to the benchmark including all the unpublished anonymous submissions. To our knowledge, RRC is the first single stage detector to achieve such result. This result not only confirms the effectiveness of RRC but also paves a new way for accuracy improvement for single stage detectors.

RRC also achieved state-of-the-art results on pedestrians and cyclist benchmark which measures IoU of 0.5. See table table \ref{table:final_compare3} and table table \ref{table:final_compare4}. Comparing to the previous published methods, we observed obvious improvements. When including all the anonymous unpublished submissions, RRC ranks the first for cyclist detection and the second for pedestrian detection. This fully justifies the effectiveness and robustness of the proposed RRC model. More qualitative results are shown in figure \ref{fig:final}.

\section{Concluding Remarks}
In this paper, we proposed a novel recurrent rolling convolution architecture to improve single stage detectors. We found RRC is able to gradually and consistently aggregate relevant contextual information among the feature maps and generate very accurate detection results. RRC achieved state-of-the-art results in all the three benchmarks in KITTI detection. To our knowledge, this is the first single stage detector to obtain such convincing results. The code is publicly available.

In the future work, we planned to investigate the memory enabled recurrent architecture in the context of object detection and quantify its impact to the detection performance. We are also interested in generalizing RRC to the task of 3D object detection and related applications.

{\small
\bibliographystyle{ieee}
\bibliography{egbib}
}

\end{document}